\documentclass{article} 
\usepackage{iclr2027_conference,times}

\usepackage{amsmath,amsfonts,bm}

\def\eqref#1{equation~\ref{#1}}

\def\1{\bm{1}}

\DeclareMathAlphabet{\mathsfit}{\encodingdefault}{\sfdefault}{m}{sl}
\SetMathAlphabet{\mathsfit}{bold}{\encodingdefault}{\sfdefault}{bx}{n}

\usepackage{hyperref}
\usepackage{url}
\usepackage{enumitem}
\usepackage{booktabs}
\usepackage{graphicx}
\usepackage{xcolor}
\usepackage{algorithm}

\usepackage{algorithmic}
\usepackage{amsthm}
\usepackage{float}
\usepackage{csquotes}
\MakeOuterQuote{"}   

\title{Does the Explanation Drift Before\\the Guarantee? Label-Free Early Warning on Evolving Graphs}

\author{Tomasz Szandala \\
Faculty of Information and Communication Technology \\
Wroclaw University of Science and Technology\\
\texttt{tomasz.szandala@pwr.edu.pl} \\
\And
Damian Raczkowski\\
independent researcher\\
\texttt{damian.raczkowski.dr@gmail.com} \\
}

\iclrfinalcopy 
\begin{document}

\maketitle

\begin{abstract}
Deployed models can fail silently as the world they describe evolves. Attribution drift has served as an early warning in tabular data; we ask whether, in graph streams, a model's own explanations can forewarn the loss of nominal conformal coverage.
Doing so requires repairing the naive drift statistic twice. A vanishing edge loses mask mass regardless of whether the model relied on it, so we condition the importance on edge presence. Evolving graphs also break edge identity: a relabelling that changes nothing shifts the identity-based statistic by 0.69 on synthetic graphs and 0.83 on real ego-networks, whereas aggregating over structural roles brings both to 0.0 and stays well-defined when the support changes.
We test six pre-registered regimes on thirty seeds each. A two-channel monitor, tracking conditional mask mass and its distribution over roles, precedes coverage collapse by 8 to 14.5 steps across four gradual regimes (node addition, removal, replacement, and edge thinning), all surviving Holm correction. A sudden knockout yields the expected zero lead, and a harmless change triggers no alarm. Leads track the length of the transitional phase (Spearman 0.79 over 150 runs). On real transaction ego-networks, the untuned mass channel forewarns staged drift by 27 steps and is the only channel, attributional or structural, that never alarms on a stationary stream.
The main limitation is that the mass channel is blind to covariate drift, which breaks the guarantee in every run; the distributional channel detects it and also rescues the rewiring regime, so we report the two channels separately rather than as a union.
\end{abstract}



\section{Introduction}
A trained model becomes less useful as the world changes. This degradation can happen silently: the labels that would reveal it may arrive late, be costly to obtain, or never arrive at all~\citep{demvsar2018detecting}. A practical monitor should therefore warn of failure \emph{before} it occurs, without relying on labels or other external warning signals.

Feature attributions provide a natural label-free signal, and prior work has used them for drift monitoring~\citep{mougan2022explanation,mougan2025explanation,haug2022change,duckworth2021using}. Two gaps remain. First, existing studies ask whether attributions change, not whether they change \emph{before} model performance deteriorates-the property that makes a warning operationally useful. Second, they focus on tabular data, where each feature occupies a fixed coordinate. Graph models instead reason over topology, and that topology itself evolves as edges appear, disappear, and rewire. For graph streams, we therefore ask: does explanation drift precede model degradation, and if so, by how much? We call this interval the \emph{lead}.

\begin{figure}[t]
\centering
\includegraphics[width=0.78\textwidth]{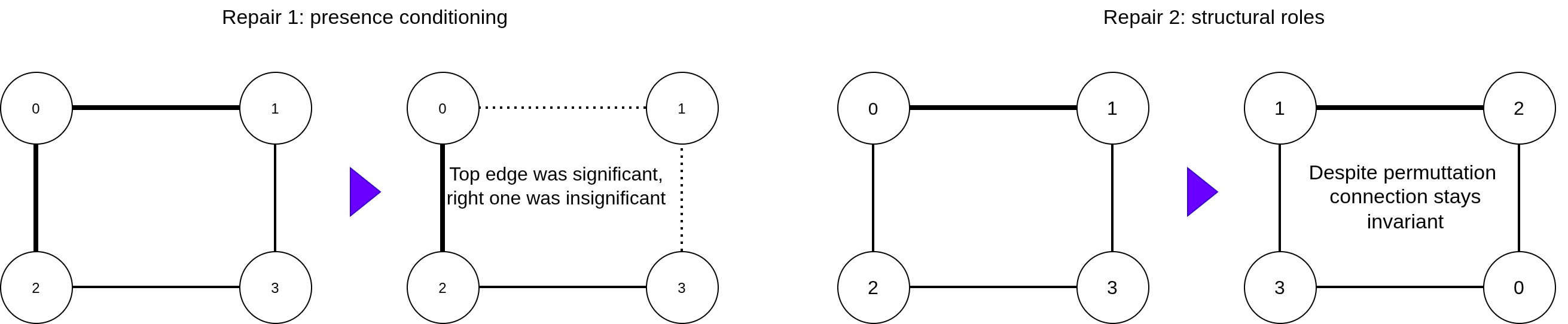}
\caption{The two repairs on four edges. (a) When an edge thins, the raw mask loses mass whether or not the model relied on it; dividing by presence isolates the model's dependence per unit of presence. (b) A relabelling (or a sampled instance) changes what coordinate $(0,1)$ \emph{names}; the role key---the unordered pair of endpoint degree bins---names the same structural object regardless.}
\label{fig:repairs}
\end{figure}

Answering this question requires two repairs to the drift statistic. The first concerns \emph{what the statistic measures}. A learned edge mask has the form $M_t=\sigma(\ell_t)\odot A_t$, so when an edge disappears, its mask mass vanishes regardless of whether the model's reliance on that edge changed. Conditioning importance on edge presence separates "the model stopped using this edge" from "this edge disappeared" (Figure~\ref{fig:repairs}a).

The second concerns \emph{what the statistic is indexed by}. Coordinate $(i,j)$ is meaningful across time only if it refers to the same object, which evolving graphs do not guarantee: nodes enter and leave, while sampled instances, such as ego-networks, reshuffle coordinates. A random relabelling that leaves the graph unchanged moves an identity-based statistic by $0.69$ on synthetic graphs and $0.83$ on real ego-networks. We therefore aggregate presence-conditioned importance over structural roles: an edge is indexed by the structural roles of its endpoints-degree bins, optionally refined by WL~\citep{shervashidze2011weisfeiler}-rather than by their identities (Figure~\ref{fig:repairs}b). The resulting statistic is invariant to relabelling and remains well-defined when graph support changes.

We evaluate warning time against the coverage of a split-conformal predictor~\citep{vovk2005algorithmic,angelopoulos2023gentle}: whether its prediction sets continue to contain the true label at the target rate, e.g., $90\%$. The mechanism behind a positive lead is simple. Prediction-level quantities can remain stable while any sufficient substructure survives, whereas attributions respond as soon as the model shifts reliance between substructures. Whether a lead exists, therefore, depends on whether the model has a fallback substructure, which can be assessed on the frozen model before deployment. Its maximum duration is constrained by the transitional phase induced by the drift. This mechanism yields predictions for six drift regimes, including two negative controls-a sudden knockout and a harmless change-which distinguish an early-warning signal from a merely faster detector or a generic novelty detector.

\textbf{Contributions}
\begin{itemize}[nosep,leftmargin=1.2em]
\item We formulate explanation drift as an \emph{early-warning} problem for graph streams and define lead time in terms of split-conformal coverage degradation. We address two failures of standard attribution-drift statistics: structural confounding via presence conditioning and unstable edge identity via role aggregation. The latter is provably permutation-invariant and reduces a relabelling-induced shift from $0.69$ to $0.0000$ at comparable sampling noise.
\item We test the proposed mechanism on six pre-registered drift regimes and a covariate control, with $30$ seeds each. Five of the six registered predictions hold; rewiring defeats the mass channel but is recovered by a post-hoc distributional channel. Lead time is strongly associated with transition length (Spearman $\rho=0.79$, $n=150$).
\item We test scaling to $50$, $70$, and $100$ nodes. Positive leads persist, negative controls remain silent, and the conformal margin increases without retuning.
\item On real transaction ego-networks, the identity failure reproduces, and injected staged drift is detected a median $27$ steps before harm onset ($15/15$ detections, $0/15$ false alarms, $p<0.001$). Among the tested attributional and structural channels, the mass channel is the only one with zero false alarms across all $15$ stationary runs.
\item We characterise the boundaries: the mass channel misses covariate drift despite coverage degradation, and no tested label-free channel forewarns an abrupt real-world market shutdown.
\end{itemize}

\section{Related Work}
\label{sec:related}

\citet{mougan2022explanation,mougan2025explanation} formalise \emph{explanation shift} and detect it from SHAP distributions, while stating plainly that detecting a shift is not predicting a performance loss; the wider monitoring literature~\citep{duckworth2021using,shanbhag2021unified,haug2022change,lopatecki2024data} likewise treats attributions as a label-free alarm, and \citet{hinder2023model} explain drift retrospectively. All of it is tabular, and none asks whether the alarm sounds \emph{before} the model fails, or by how much.

Label-free drift detection is mature: DDM~\citep{gama2004learning}, ADWIN~\citep{bifet2007learning}, Page--Hinkley~\citep{page1954continuous} and the model-based STUDD~\citep{cerqueira2023studd} raise alarms from inputs, outputs or a student--teacher discrepancy. We adopt this machinery unchanged (\S\ref{sec:detector}) and point it at graph attributions.

Graph explanations have been studied for quality and stability. PGExplainer~\citep{luo2020parameterized} is inductive---train once, then track masks cheaply over time---and \citet{agarwal2023evaluating} find it the steadiest of the field. Counterfactual explainers such as CF-GNNExplainer~\citep{lucic2022cf} offer an alternative per-edge signal. Throughout, stability is a static virtue to be maximised, never turned into a timed warning.

Conformal prediction under shift repairs the guarantee after a known kind of shift: weighted calibration for covariate shift~\citep{tibshirani2019conformal}, corrections for label shift~\citep{podkopaev2021distribution}, online re-calibration~\citep{gibbs2021adaptive}. We use the guarantee's failure as the harm event and ask whether a label-free signal precedes it---a complementary question, and one their corrections could eventually act upon.

\section{Method}
\label{sec:method}

We monitor a frozen graph classifier $f$ together with a frozen edge explainer $g$ (PGExplainer-style~\citep{luo2020parameterized}) on a stream of graph batches of size $B$. The deployment scenario is a static model kept in service while its inputs evolve---the common case for a monitor---rather than a temporal GNN that updates its own memory~\citep{rossi2020temporal,xu2020inductive}; the latter would need a monitor of its own, since its explanations change by design. A labelled reference window $[0,R)$ is split in two: $f$ and $g$ are trained on $[0,R_0)$ ($g$ against $f$'s predicted labels), and $[R_0,R)$ is held out for calibration. From $R$ onwards, monitoring is label-free. At step $t$ the explainer yields a mask $M_t=\sigma(\ell_t)\odot A_t$.

\subsection{Signals: from raw masks to roles}
\label{sec:signals}

Averaging $M_t$ over the batch and vectorising the upper triangle gives an importance vector $\mathbf{a}_t$ indexed by node pairs. Since most edges are irrelevant, we track the loss of mass on the reference top-$k$ set $\mathcal{K}$ of a chosen importance vector $\mathbf{v}$:
\begin{equation}
D_{\text{focus}}(t) =
\frac{\sum_{i\in\mathcal{K}} v_{\text{ref},i} - \sum_{i\in\mathcal{K}} v_{t,i}}
{\sum_{i\in\mathcal{K}} v_{\text{ref},i}}.
\label{eq:dfocus}
\end{equation}
Eq.~\ref{eq:dfocus} is one-directional by design: it registers \emph{loss} of reliance on what the model used at reference time, not new reliance on structures that were unimportant then. Redistribution is the job of the second channel introduced below.

\textbf{Repair 1: presence conditioning.}
With $\mathbf{v}=\mathbf{a}$, Eq.~\ref{eq:dfocus} is confounded: the mask contains $A_t$ as a factor, so a vanishing edge loses mass regardless of the model's reliance. Let $\varphi_{t,i}$ be the presence frequency of edge $i$ in the batch. The conditioned importance
\begin{equation}
c_{t,i} = \frac{a_{t,i}}{\max(\varphi_{t,i},\,\varepsilon)}
\label{eq:cond}
\end{equation}
asks how important the edge is \emph{given that it is present} ($\mathcal{K}$: reference presence above $0.3$). This one division separates measuring the model from measuring the graph: when a non-predictive decoy motif degrades alongside the predictive one, the conditioned signal ranks the predictive edges first (precision $0.62$) while the raw signal and a structural baseline score $0.00$ (Appendix~\ref{app:decoy}).

\textbf{Repair 2: role aggregation.}
Both variants are indexed by edge identity, which evolving graphs do not preserve. We aggregate by structure instead: each node gets a colour, and the \emph{role} of an edge is the unordered pair of endpoint colours---a key defined by the graph alone. Roles are defined on the \emph{support}, the set of node pairs actually present in a batch, so they remain meaningful when that set changes. For batch $t$, let $m_t(\rho)$ be the per-graph total mask mass on edges of role $\rho$, averaged over the batch, and $\varphi_t(\rho)$ the per-graph count of such edges, likewise averaged; both count present edges only. Conditioning transfers:
\begin{equation}
c_t(\rho) = \frac{m_t(\rho)}{\max(\varphi_\bullet(\rho),\,\varepsilon_\rho)},
\label{eq:condrole}
\end{equation}
where $\bullet$ is either the current step or the reference window. The choice matters. The \emph{current} frequency is invariant even to support change, but it divides out exactly the loss under watch; the \emph{reference} frequency keeps invariance to identity and sensitivity to support loss. Our primary channel, the \emph{mass channel} $D_{\text{role}}$, uses the reference denominator; a pre-registered ablation $D_{\text{role}}^{\text{cur}}$ confirms the trade-off under node removal ($+6.5$ against $+13.0$ steps of lead, Appendix~\ref{app:ablations}). A second channel, the \emph{distributional channel} $D_{\text{role}}^{L1}$, normalises the conditioned profile $c_t(\cdot)$ to sum to one over roles and takes the total variation distance to the reference profile; it sees \emph{redistribution} of mass across roles even when the total is conserved.

\textbf{Why the role profile survives relabelling.}
\label{par:inv}
Apply an arbitrary node permutation $\pi_b$ to each graph $b$ of a batch. Colours are isomorphism-invariant, so a permuted node keeps its colour and every present edge contributes to the same role key as before; the per-role sums are therefore identical, and if encoder and explainer are permutation-equivariant the whole profile $(m_t,\varphi_t)$ is unchanged, and with it $D_{\text{role}}$, $D_{\text{role}}^{\text{cur}}$ and $D_{\text{role}}^{L1}$. The identity vector $\mathbf{a}_t$ enjoys no such protection: it is permuted along with the nodes, so a focus set fixed on the reference window reads different entries after relabelling and the statistic moves although nothing about the data has. Section~\ref{sec:invariance}
measures both.


\subsection{The protected quantity: conformal coverage}
\label{sec:conformal}
We protect the guarantee of a split-conformal predictor~\citep{vovk2005algorithmic,angelopoulos2023gentle}. Nonconformity scores $s=1-p_f(y\,|\,G)$ on the calibration slice $[R_0,R)$ fix the threshold $\hat q$ (the $\lceil(n_{\text{cal}}{+}1)(1{-}\alpha)\rceil$-th smallest of the $n_{\text{cal}}$ calibration scores); prediction sets $\{y: s(y)\le\hat q\}$ then contain the truth at rate $\ge 1-\alpha$ on exchangeable data. We track the coverage gap $g(t)=(1-\alpha)-\mathrm{cov}(t)$. Its calibrated change point is the harm onset $t_{\text{cov}}$, and the lead of a signal with onset $t_{\text{attr}}$ is $L=t_{\text{cov}}-t_{\text{attr}}$. Coverage needs labels, so it is the \emph{evaluation} target only; the monitor stays label-free after calibration. It is a stricter clock than accuracy, since confidence erodes before the argmax flips.

Coverage has a pitfall. A saturated reference model piles calibration scores at zero, so $\hat q\approx0$ and coverage becomes a hair trigger that collapses on harmless changes (a gap of up to $0.53$ in our setting). In the synthetic generator we avoid it by grading reference difficulty---a per-instance evidence strength $u\sim\mathcal{U}(0.5,1)$---which gives $\hat q$ a genuine margin (median $0.018$, calibration coverage $0.901$); this is a property of the benchmark, not an intervention the monitor performs. The problem is specific to small graphs: growing them from $25$ to $100$ nodes with everything else fixed raises $\hat q$ from $0.018$ to $0.132$ on its own (\S\ref{sec:scale}).

\subsection{Detector, coupling, and controls}
\label{sec:detector}
Every signal runs through the same calibrated Page--Hinkley detector~\citep{page1954continuous} (robust standardisation on $[R_0,R)$, tolerance $\delta$, threshold $\lambda$; quantised series get a $0.5/B$ floor). Beyond timing, we report the \emph{coupling} $\Psi$, the Spearman correlation between the smoothed signal and the coverage gap over the monitored window. The lead is not an artefact of these choices: four detectors and a gradient-based explainer reproduce it, whereas a counterfactual edge-deletion score does not, since it tracks prediction flips and so moves on the very clock the monitor is trying to beat (Appendix~\ref{app:ablations}). On real data we add a \emph{prediction-prior} channel---drift of the model's predicted class rate on natural batches---which needs no labels and tracks label shift rather than reasoning shift.

\subsection{When can the explanation move first?}
\label{sec:theory}


\textbf{The condition: a plan B.}
Suppose a positive graph can be recognised from either of two substructures, $A$ or $B$, and drift erodes $A$ first. The prediction does not care which one it uses: it stays correct as long as \emph{either} survives and fails only when the last one goes. The explanation does care: as $A$ becomes less useful, the mask shifts importance from $A$ to $B$ straight away. The explanation therefore changes while the prediction holds, and the time between the two is the lead.

For a lead to exist, the model must have a plan B. We measure this on the frozen model: delete $A$ from the reference-window graphs and record the accuracy that survives, which we call the \emph{survivor capability}. Seed-to-seed training variation supplies variation in it for free. Of thirty models trained on the same generator, eighteen clear a pre-registered threshold of $0.96$ and twelve do not, and that split separates a median lead of $+20.5$ steps from one of $-3.5$ (\S\ref{sec:regimes-results}). Two further checks confirm that the mechanism is literally what happens: accuracy falls $0$--$1$ steps after the last survivor vanishes, and the mask moves $1$--$2$ steps after $A$ stops being useful, never before ($0/10$). $A$ stops being useful a median $33$ steps before accuracy falls: the lead is the time the model spends living on its plan B.

This check needs labels and knowledge of which substructure will erode, so it is a design-time property a model's owner can establish on training data, not a run-time test. The experiments establish the implication: where redundancy is absent, the lead is absent too, and the monitor's silence is then correct rather than blind.

\textbf{The bound: the transitional phase.}
A plan B decides whether a lead exists. How long it can be is decided by the drift, not the monitor: the lead is bounded by the \emph{transitional phase}, the stretch during which the model has already moved onto $B$ while its guarantee still holds. This fixes a prediction for each regime of Table~\ref{tab:regimes}. Drifts that erode $A$ gradually leave a phase, so $\textit{lead}>0$. A sudden drift that destroys $A$ and $B$ together leaves none, so $\textit{lead}\le0$. A change that touches neither must leave the monitor silent. The last two are controls: a lead under knockout would mean we had merely built a faster detector, and an alarm on the null that we had built a novelty detector. The mechanism also says where the monitor is blind: harm arriving through something the explanation does not track (background density, say) leaves the model's reasoning---and hence the mask---in place. The \textsc{covariate} control measures that blind spot; it is a harmful regime that the attribution channel is predicted to miss.

\begin{table}[t]
\centering
\small
\begin{tabular}{llll}
\toprule
Regime & Stage-one mechanism & Support & Predicted lead \\
\midrule
\textsc{thinning} & $A$'s density decays in place & fixed & $+$ \\
\textsc{knockout} & both motifs collapse at once & fixed & $0$ (no transitional phase) \\
\textsc{add} & $A$'s edges rewired through new nodes $C$ & grows & $+$ \\
\textsc{remove} & $A$'s nodes retired entirely & shrinks & $+$ \\
\textsc{mix} & removal of $A$ plus non-predictive $C$ traffic & both & $+$ \\
\textsc{null} & $A,B$ untouched; background-level traffic on $C$ & grows & silent \\
\textsc{covariate} & $A,B$ untouched; background density grows & fixed & harm, no signal \\
\bottomrule
\end{tabular}
\caption{Six drift regimes with outcomes registered before the campaign, plus the covariate control that measures the predicted blind spot. A further pre-registered prediction: the current-frequency denominator underperforms under \textsc{remove}. Stage one occupies drift progress $s\in[0,0.4]$; stage two ($s\in[0.6,1]$) degrades the survivor $B$.}
\label{tab:regimes}
\end{table}

\section{Experiments}
\label{sec:results}

We test the monitor on two substrates, ordered by how badly identity fails: a synthetic stream with moving support (the six registered regimes and the covariate control) and sampled real ego-networks. A third campaign varies graph size. Shared synthetic setup: GIN classifier, $T=130$ steps, batch $64$, reference window $45$, drift over $[55,115]$; $35\%$ of positives carry $B$ alone (Appendix~\ref{app:repro2}).

\subsection{Moving support: outcomes against the pre-registered predictions}
\label{sec:regimes-results}
\begin{figure}[t]
\centering
\includegraphics[width=0.55\textwidth]{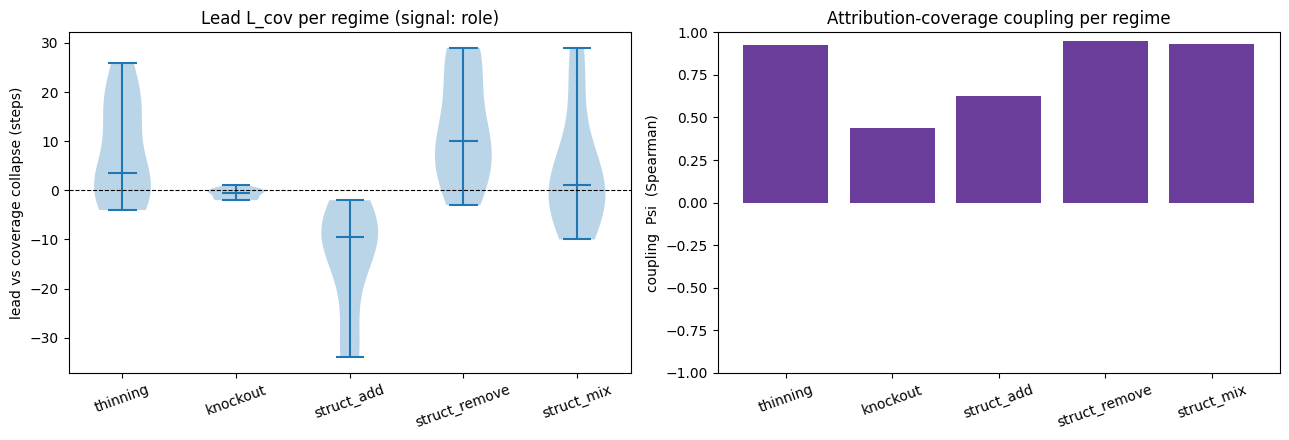}
\caption{Per-regime leads against coverage collapse (left) and attribution--coverage coupling $\Psi$ (right), $30$ seeds per regime.}
\label{fig:regimes}
\end{figure}

The regime campaign runs Table~\ref{tab:regimes} on the coverage clock: $30$ seeds per regime, $\alpha=0.1$, evidence jitter as in \S\ref{sec:conformal} (preconditions in Appendix~\ref{app:repro2}). Results: Table~\ref{tab:main}, Figure~\ref{fig:regimes}.

\begin{table}[t]
\centering
\footnotesize
\setlength{\tabcolsep}{4pt}
\begin{tabular}{llrrrrrl}
\toprule
Regime & Pred. & $L$ (med.) & IQR & frac $L>0$ & Holm $p$ & $\Psi$ & Verdict \\
\midrule
\textsc{thinning} & $+$ & $+9.0$ & $[-3.2,\,21.0]$ & $0.70$ & $0.005$ & $0.92$ & confirmed \\
\textsc{knockout} & $0$ & $-1.0$ & $[-2.0,\,-0.2]$ & $0.03$ & - & $0.35$ & confirmed (control 1) \\
\textsc{add} & $+$ & $-14.0$ & $[-20.8,\,-6.0]$ & $0.07$ & $1.000$ & $0.66$ & failed; \S\ref{sec:twochannel} \\
\textsc{remove} & $+$ & $+13.0$ & $[5.5,\,30.0]$ & $0.93$ & $<10^{-4}$ & $0.92$ & confirmed \\
\textsc{mix} & $+$ & $+8.0$ & $[-2.0,\,22.8]$ & $0.70$ & $0.002$ & $0.91$ & confirmed \\
\textsc{null} & silent & - & - & $0/30$ alarms & - & $-0.05$ & confirmed (control 2) \\
\textsc{covariate} & blind & - & - & $0/30$ alarms & - & $-0.88$ & blind spot, measured \\
\bottomrule
\end{tabular}
\caption{Mass channel $D_{\text{role}}$ against coverage collapse; one-sided Wilcoxon tests of $L>0$, Holm-corrected over the four regimes predicting a lead; $30$ seeds per regime; detection rate $1.00$ in every harmful regime. Under \textsc{covariate} the guarantee collapses in $30/30$ runs while the signal stays silent in all $30$; on \textsc{null}, coverage itself wobbled in $7/30$ runs, the signal in none.}
\label{tab:main}
\end{table}

Five rows behave as registered. \textsc{remove} is the regime the role statistic exists for: coordinates die, yet the signal leads by $+13.0$ steps ($p<10^{-4}$). \textsc{knockout} alarms with no lead, where a generically faster detector would lead. \textsc{null} yields no alarm in $30$ runs. Under \textsc{covariate} the model is harmed through a channel the explanation does not carry, and the monitor correctly says nothing while the guarantee fails in every run: the blind spot is a prediction of the mechanism, not a gap in the measurement. The registered denominator ablation also held (\S\ref{sec:signals}). Coupling $\Psi$ is high in every harmful regime, near zero on \textsc{null}, and strongly negative under \textsc{covariate}.

\textbf{The condition predicts which runs carry a lead.}
With the generator fixed, ordinary training variation leaves $18$ of $30$ models above the pre-registered survivor-capability threshold of $0.96$ and $12$ below. The threshold is measured on the frozen model before any drift, and the split is sharp. Above it, \textsc{remove} leads by $+29.0$ steps with every run positive, \textsc{thinning} by $+20.5$, \textsc{mix} by $+14.0$. Below it, the same regimes give a median lead of $-3.5$ with fewer than half the runs positive: no plan B, no forewarning. The unconditional table remains the headline; the split is evidence about the mechanism, not a better operating point---but a practitioner can measure the redundancy of a model before deploying it.

\textbf{Structural baselines see the graph; the attribution sees the model.}
We place every channel at its own zero-false-alarm threshold, calibrated on the harmless \textsc{null} regime, and compare them (Table~\ref{tab:opbase}, Appendix~\ref{app:opbase}). The structural baselines lead comparably---degree MMD by $+8.0$ steps pooled against the mass channel's $+3.0$---and, unlike the mass channel, they alarm under \textsc{covariate} ($100\%$ and $97\%$ of runs), where coverage does collapse. Those alarms are true positives; the mass channel's silence there is its blind spot, not a virtue. What the attribution channel provides instead is a more model-specific signal: its alarm is associated with a change in the model's attributed reliance, which a structural signal cannot distinguish from a density change. At the shared default threshold degree MMD also false-alarms on $87\%$ of harmless \textsc{null} runs, and on the stationary real controls of \S\ref{sec:injected} both structural baselines false-alarm where the mass channel does not. Predictive uncertainty never leads at any threshold. The two kinds of signal are complements, not rivals (\S\ref{sec:disc}).

\textbf{Invariance, measured directly.}
\label{sec:invariance}
Permuting the node labels of every graph in a reference batch leaves the data unchanged, so a correct statistic must not move; over $20$ trials the identity statistic moves by $0.69$ and the role statistic by $0.0000$. Bootstrap sampling noise is comparable for both ($0.069$ vs $0.054$): the decisive difference is identity, not variance.

\subsection{The \textsc{add} failure and the two-channel repair}
\label{sec:twochannel}
Under \textsc{add} the mass channel wakes after the harm ($-14.0$): rewired $A$--$C$ edges fall into the same degree-bin role as the vanished $A$--$A$ edges, so the per-role balance hides the migration. Coarseness is not the cause: a coarser colouring ($6$ roles) and a finer one ($21$ roles) leave the \textsc{add} lead at $-14.0$ and $-18.0$, while every other regime shifts by a few steps at most and the mass channel still false-alarms on $0/30$ harmless runs. The failure is structural, not a tuning accident.

What recovers it is the distributional channel, which tracks the \emph{shape} of the role profile rather than its balance: mass for \emph{loss} (it carried \textsc{remove}), distribution for \emph{redistribution} (it carries \textsc{add}). The union monitor alarms at the first onset of either; the $L1$ threshold is chosen on twenty \textsc{null} seeds and evaluated on ten held-out seeds, on which it false-alarms once (Table~\ref{tab:union}). This repair is post hoc, so \textsc{add} is a recovered failure, not a confirmed prediction. It is also paid for in specificity: the distributional channel alarms on $29/30$ covariate runs where the mass channel alarms on none (\S\ref{sec:disc}).

\begin{table}[t]
\centering
\small
\begin{tabular}{lrrrr}
\toprule
Regime & $L_{\text{union}}$ (median) & IQR & frac $L>0$ & Holm $p$ \\
\midrule
\textsc{thinning} & $+14.5$ & $[5.5,\,23.5]$ & $0.87$ & $<10^{-4}$ \\
\textsc{knockout} & $0.0$ & $[-1.0,\,0.0]$ & $0.13$ & - \\
\textsc{add} & $+8.5$ & $[0.5,\,25.2]$ & $0.73$ & $0.0005$ \\
\textsc{remove} & $+13.0$ & $[5.5,\,30.0]$ & $0.93$ & $<10^{-4}$ \\
\textsc{mix} & $+8.0$ & $[-2.0,\,22.8]$ & $0.70$ & $0.0007$ \\
\midrule
\multicolumn{5}{l}{\emph{Alarms on the controls, by channel:}}\\
\multicolumn{5}{l}{\quad\textsc{null}: $D_{\text{role}}$ $0/30$; $D_{\text{role}}^{L1}$ and union $0/20$ on the tuning seeds, $1/10$ on the held-out seeds.}\\
\multicolumn{5}{l}{\quad\textsc{covariate}: $D_{\text{role}}$ $0/30$, $D_{\text{role}}^{L1}$ $29/30$, union $29/30$.}\\
\bottomrule
\end{tabular}
\caption{Two-channel monitor: all four transitional regimes carry leads that survive a Holm correction ($p\le0.001$ throughout), and the knockout control still yields none. The mass channel never false-alarms on either control but wakes late on \textsc{add}; the union catches \textsc{add} and inherits the distributional channel's sensitivity to background density.}
\label{tab:union}
\end{table}

\subsection{The lead is ordered by the transitional phase}
\label{sec:gradient}
The mechanism predicted signs; the data return an ordering. Each harmful run has its own transition length $t_{\text{cov}}-t_{\text{onset}}$ and its own lead, and pooled over all $150$ runs the two correlate at Spearman $0.79$ ($95\%$ bootstrap CI $[0.73,\,0.84]$; Figure~\ref{fig:gradient}, Appendix~\ref{app:gradient-fig}). This is not an artefact of pooling regimes with different drift schedules: within each regime, where only the seed varies, the correlation is $0.66$ (\textsc{knockout}), $0.73$ (\textsc{add}), $0.79$ (\textsc{thinning}), $0.93$ (\textsc{remove}) and $0.95$ (\textsc{mix}). The bound $L\le t_{\text{cov}}-t_{\text{onset}}$ holds in every run. Across regime medians the ordering has one exception, and it is the expected one: \textsc{add} grants the longest transition ($44$ steps) but only the third largest lead, because it is carried by the post-hoc distributional channel, which was never tuned for it (per-regime medians in Appendix~\ref{app:gradient-fig}).

\subsection{Scale: larger graphs, and a margin that widens on its own}
\label{sec:scale}
We repeat all seven regimes on graphs of $50$, $70$ and $100$ nodes, $10$ seeds each, holding motif size and mean background degree fixed so that a larger graph progressively \emph{dilutes} the predictive signal---the realistic case, since real graphs are large and motifs are local. Three things hold (Figure~\ref{fig:scale}, Appendix~\ref{app:scale-fig}). The role space stays stable, with a median of $10$ roles at every size. The leads persist: under \textsc{remove}, $+13.0$, $+20.5$, $+15.5$, $+17.5$ steps at $n=25,50,70,100$; under \textsc{mix}, $+8.0$, $+16.5$, $+11.5$, $+16.5$. And the controls stay silent at every size ($0/10$ alarms on \textsc{null} and on \textsc{covariate}, the latter while the guarantee collapses in $10/10$), so the detector threshold transfers across densities. Two by-products: the conformal margin widens with size unaided ($\hat q$ median $0.018$, $0.057$, $0.111$, $0.132$), and the \textsc{null} coverage wobble at $n=25$ disappears at $n\ge50$. The \textsc{add} deficit shrinks with scale ($-14.0$, $-7.0$, $-2.0$, $-7.5$) without turning positive.

\subsection{Real ego-networks}
\label{sec:elliptic}
\label{sec:injected}
\begin{figure}[t]
\centering
\includegraphics[width=0.56\textwidth]{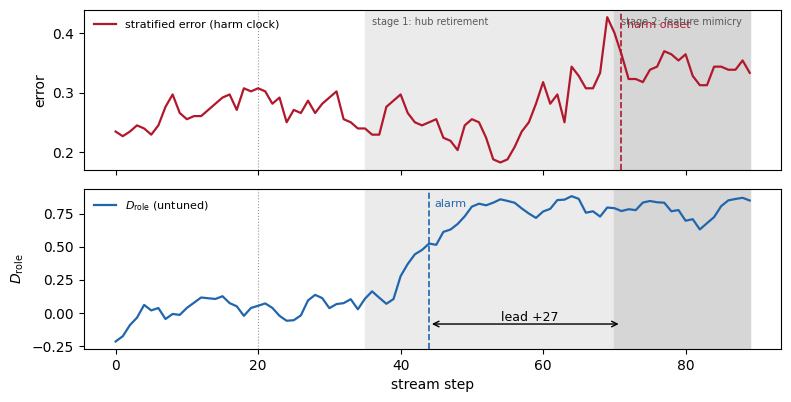}
\caption{Injected staged drift on Elliptic ego-networks, one repetition. Top: the harm clock (stratified error); bottom: the untuned mass channel $D_{\text{role}}$. Stage one retires hub nodes, stage two makes illicit instances mimic licit ones; the attribution alarm fires in stage one, $27$ steps before the harm onset.}
\label{fig:injected}
\end{figure}
The organic Elliptic event (\S\ref{sec:elliptic}) is abrupt, the case the mechanism excludes, so we also inject staged drift into Elliptic ego-networks to test it where its assumption holds: instances are drawn i.i.d.\ from a single pooled calibration urn (stationary by construction; a no-injection control confirms it), stage one progressively retires the highest-degree non-seed node of each ego-network, and stage two makes illicit instances \emph{mimic} licit ones by transplanting licit node features---the adversarial adaptation any fraud model eventually faces. The prediction, fixed in advance: the attribution moves at stage one, harm arrives at stage two, and the lead is positive.

It is (Figure~\ref{fig:injected}). The mass channel alone, at its default threshold and with \emph{no} tuning on this substrate, fires in $15/15$ repetitions during stage one and leads the harm clock (which fires in $14/15$) by a median $+27.0$ steps (bootstrap $95\%$ CI $[23.0,\,30.0]$; $13/14$ positive; Wilcoxon $p=0.0006$), with $0/15$ false alarms on the stationary controls. Because the controls are i.i.d.\ draws from one urn, every alarm on them is false by construction---the sharpest specificity test available, and the mass channel is the only channel that passes it. At the same default threshold (not at their own transferred zero-false-alarm thresholds; see \S\ref{sec:disc}), degree MMD alarms on $2/15$ stationary controls, embedding MMD on $3/15$ and the distributional channel on $3/15$; their leads ($+32.0$, $+30.0$, $+30.0$) are a few steps longer and are bought with those alarms. The union monitor, transferred from the synthetic campaign without re-tuning, gives $+30.0$ at $2/15$.

The harm clock here is stratified accuracy, because the illicit-class conformal threshold saturates ($\hat q_1\ge0.99$ in $14/15$ repetitions): the class is never excluded, the guarantee holds vacuously, and prediction sets inflate ($11/15$). A monitor must track informativeness alongside coverage (Appendix~\ref{app:coverage-levels}).

The hardest case: instances that are \emph{sampled} subgraphs, with no coordinate system connecting one to the next. We stream the Elliptic Bitcoin graph by its native time steps; each instance is a two-hop ego-network around a labelled transaction. Protocol, fixed in advance: train on steps $1$--$12$, calibrate on $13$--$25$, monitor label-free from $26$; the darknet-market shutdown at step $43$ is used for reporting only. Fifteen repetitions; $64$ ego-networks per step, stratified by true class to a fixed composition so that no reasoning statistic tracks the class mix. This stratification uses labels and is an evaluation design, not a deployable step (\S\ref{sec:disc}; Appendix~\ref{app:repro2}).

\textbf{The identity failure reproduces on real data.}
An independent random node permutation of every ego-network in a reference batch moves the identity statistic by $0.831$ (median over trials and seeds) and the role statistic by $0.0000$, at comparable step-to-step noise ($0.22$ vs $0.26$). Here this is a validity condition, not an ablation: rows of different instances denote different transactions, so the identity statistic never measured one object over time.


\textbf{A real-world knockout, and an honest miss.}
No label-free channel forewarns or co-fires with the shutdown: the role channels and the prediction-prior channel all stay silent ($0/15$ each), and the coupling is \emph{negative} ($\Psi$ median $-0.23$ for the mass channel, $-0.33$ and $-0.53$ for the distributional and identity channels): mask mass on the reference roles rises while the illicit guarantee collapses. This is the configuration the mechanism excludes, in the setting known to require label-shift-specific conformal corrections~\citep{podkopaev2021distribution}. An exogenous seizure on a date grants no transitional phase---the field analogue of \textsc{knockout}---and the new illicit transactions \emph{look different}, a concept shift on the minority class, so the model misses them without re-basing its reasoning. The shutdown thus lands in the same measured blind-spot category as the covariate control, now on real data; we report the miss rather than reframe it.

\section{Discussion}
\label{sec:disc}

\paragraph{There is no free statistic.}
Each invariance comes with a loss of sensitivity. Role aggregation removes dependence on node identity, making the statistic robust to relabelling and sampling, but it also makes structurally indistinguishable motifs impossible to separate. Likewise, using the current role frequency achieves invariance to support change but also normalises away part of the support loss under \textsc{remove}, reducing the lead. Finally, the distributional channel that recovers \textsc{add} is less specific than the mass channel: it alarms on $29/30$ covariate runs versus $0/30$ for the mass channel. These trade-offs suggest complementary operating points rather than a universally preferable statistic. A deployment concerned with any harmful drift may combine attributional and structural signals, whereas one interested specifically in changes associated with the model's attributed reliance may retain the mass channel and pair it with a structural detector for density-driven shifts. Distinguishing covariate from concept shift more finely would require a benchmark designed explicitly for that purpose.

\paragraph{Coverage is a demanding target.}
Three observations extend beyond graph monitoring. First, when the reference model is saturated, empirical coverage becomes highly sensitive to small changes, turning it into a poor harm clock. Second, under class imbalance, marginal coverage can improve even while minority-class coverage collapses. Third, class-conditional calibration can itself become uninformative on a difficult minority class when the conformal threshold saturates, and prediction sets cease to exclude that class. These cases show that the protected quantity is not merely an evaluation detail: it must be chosen and checked for degeneracy in each deployment.

\paragraph{Limitations.}
The relation between transition length and lead time is correlational rather than a fitted predictive law. The survivor condition, which determines whether a positive lead is possible, requires labels and prior knowledge of which substructure will erode; developing a label-free proxy for this condition remains an important open problem. \textsc{add} is a recovered failure rather than a confirmed prediction, and its recovery relies on a post-hoc distributional channel. The phenomenon also depends on the explainer class: continuous attributions such as mask-based and gradient-based scores exhibit a lead, whereas the counterfactual edge-deletion score does not because it changes on approximately the same clock as the prediction itself (Appendix~\ref{app:ablations}). Role granularity is another practical constraint: WL depth $1$ fragments the $25$-node graphs into thousands of roles, at which point detection degrades sharply (Appendix~\ref{app:repro2}). Our statistics operate on batches, so continuous-time streams would require an additional windowing scheme.

On Elliptic, evaluation batches are stratified by true class to isolate attribution drift from class-prior drift. This stratification is an evaluation device rather than a deployable procedure; on natural batches, the two effects coexist, with the prediction-prior channel measuring the latter. We do not report the reasoning channels on unstratified batches. In \S\ref{sec:injected}, structural baselines are evaluated at the shared default threshold rather than at separately transferred operating points, so their stationary-control alarm rates should be interpreted as indicative rather than definitive. Finally, the only organic harmful shift observed in Elliptic is abrupt, precisely the case excluded by the proposed mechanism. Early warning of \emph{naturally occurring gradual harm} on real graph streams, thus, remains untested. Our synthetic scaling experiments also stop at $100$ nodes.
\section{Conclusion}
On graph streams, a model's explanation drifts before its guarantee fails---once the drift statistic is repaired twice, and only where the drift grants time. Six pre-registered regimes bear this out on graphs of $25$ to $100$ nodes, with leads ordered by the length of the transitional phase and both negative controls silent. On real ego-networks the same machinery forewarns injected drift by $+27$ steps without a false alarm on stationary controls; before an abrupt label shift it stays silent, the mechanism's excluded case. The claim is bounded: a label-free witness of guarantee loss that forewarns where a transitional phase exists, says in advance where it cannot, and tells reasoning change from graph change.

\newpage
\bibliography{bibliography}
\bibliographystyle{iclr2027_conference}

\newpage
\section*{Use of AI}
Claude (Anthropic; Opus 5) was used for language polishing, for condensing the manuscript to the page limit, and preparing the anonymised supplementary Colab notebook (improving readability and ensuring anonymity, all results are consistent).

\appendix
\section{Reproducibility}
\label{app:repro2}

Anonymised Colab notebooks (synthetic campaign, scale campaign, Elliptic) are included as supplementary material with this submission.

\subsection{Generator}
Stream: $T=130$ steps, batch $64$, $n=25$ nodes, motif size $K=7$ (blocks $A=[0,7)$, $B=[7,14)$, $C=[14,21)$), background edge probability $0.06$ (covariate control ramps to $0.10$), motif densities $p_{\text{hi}}=0.95$, $p_{\text{lo}}=0.10$, reference window $[0,45)$, drift over $[55,115]$; $35\%$ of positives carry $B$ only, so no label depends on $A$ and the survivor capability clears its threshold in the median run. Regime campaign adds evidence jitter: per positive graph, $u\sim\mathcal{U}(0.5,1)$, $p_{\text{eff}}=p_{\text{lo}}+u\,(p-p_{\text{lo}})$, where $p$ is the motif's scheduled density at that step ($p_{\text{hi}}$ before drift). Stage one occupies $s\in[0,0.4]$, stage two $s\in[0.6,1]$. Regime mechanics: \textsc{thinning} interpolates $p_A$ downwards; \textsc{knockout} sets both motifs to $p_{\text{lo}}$ for $s>0$; \textsc{add} rewires each intra-$A$ edge with probability $q(s)$ through a random node of $C$; \textsc{remove} deactivates all $A$-nodes (edges and features) with per-graph probability $q(s)$; \textsc{mix} combines removal with background-level $C$ traffic in both classes; \textsc{null} leaves $A,B$ untouched and ramps background-level $C$ traffic; the \textsc{covariate} control leaves $A,B$ untouched and ramps the global background density from $0.06$ to $0.10$.

Preconditions over all $210$ regime-campaign runs: reference accuracy $0.993$, survivor capability $0.798$ ($A$ alone) and $0.964$ ($B$ alone), mask selectivity $3.77\times$ the uniform share, calibration coverage $0.901$.

\subsection{Model, explainer, signals, detector}
Classifier: GIN with sum readout, trained on $[0,R_0)$; fixed-support campaign uses $R_0=R=40$ with the accuracy clock, the regime campaign uses $R_0=22$ with conformal calibration on $[22,45)$; $200$ epochs. Explainer: PGExplainer-style, trained on the frozen classifier's predictions on the training slice. Focus sets: top $42$ edges (identity signals; conditioned variant restricted to reference presence $>0.3$, floor $\varepsilon=0.05$) and top $6$ roles (frequency $\ge1$ edge per graph, floor $\varepsilon_\rho=0.5$). Roles: degree bins with edges $(1.5,3.5,6.5)$, giving a median of $10$ roles in the reference window; WL depth $h=0$ in main runs. Granularity is ablated by the bin edges rather than by WL depth: at $h=1$ the colouring fragments into a median of $6682$ roles on $25$-node graphs, per-role frequencies become noise and the detection rate collapses to $0.2$--$0.4$, so granularity is ablated by bin width instead (\S\ref{sec:twochannel}). Conformal: $\alpha=0.1$; $\hat q$ is the $\lceil(n{+}1)(1{-}\alpha)\rceil$-th smallest calibration score (order statistic, not an interpolated quantile). Detector: Page--Hinkley on smoothed, robustly standardised series ($\delta=1.5$, $\lambda=16$, smoothing $3$); error and coverage-gap series receive a $0.5/B$ standard-deviation floor. Two-channel monitor: union of $D_{\text{role}}$ at $\lambda=16$ and $D_{\text{role}}^{L1}$ at $\lambda=24$, the latter the smallest grid value with zero false alarms on the twenty \textsc{null} seeds outside the held-out block (seeds $0$-$9$ and $20$-$29$), evaluated on held-out seeds $10$-$19$ ($1/10$).

\subsection{Baselines and the union grid}
Degree-distribution MMD: RBF-kernel MMD between the degree multiset of the current batch and the reference window (median-heuristic bandwidth); embedding MMD: the same on frozen-encoder graph embeddings; feature MMD: on raw node features; predictive uncertainty: mean entropy of the frozen classifier. Each baseline's threshold is the smallest grid value with zero false alarms on its own controls. Union monitor grid for $\lambda_{L1}$: $\{16, 20, 24, 30, 40\}$; the smallest value with zero calibration false alarms was $\lambda_{L1}=24$, validated held-out ($1/10$). For the operating-point comparison of Table~\ref{tab:opbase} every channel is swept over $\lambda\in\{6,8,10,12,16,20,24,30,40,56,80\}$ and placed at its own smallest zero-false-alarm value.

\subsection{Elliptic protocol}
Two-hop ego-networks capped at $30$ nodes (BFS order), $165$ standardised node features (statistics fitted on training steps), GIN as in the synthetic campaign; $15$ organic repetitions, and $15$ injected-drift repetitions with $15$ stationary controls. Evaluation batches: $64$ ego-networks at \emph{fixed} class composition (up to $32{+}32$) -- a deliberate design choice that isolates drift of the model's reasoning from drift of the class prior, which is measured separately by the prediction-prior channel on natural batches drawn from all nodes (labelled or not; predictions need no labels). A third, parallel batch draws \emph{labelled} nodes at the natural $\approx90{:}10$ composition and is used only for the marginal-coverage comparison of \S\ref{sec:elliptic}: on a stratified batch a "marginal" quantity is in effect class-conditional, and the comparison would be vacuous. Class-conditional conformal: per-class $\hat q$ on steps $13$-$25$; the protected series is the illicit-class coverage gap with a standard-deviation floor of $0.5/n_{\text{illicit}}$. The permutation exhibit applies $10$ independent per-graph relabellings at a mid-calibration step and reports $|D(\text{perm})-D(\text{orig})|$. Injected-drift protocol: pseudo-stream of $90$ steps drawn i.i.d.\ from a single pooled calibration urn ($32{+}32$ per batch); onset at step $35$, full strength at $70$; stage one retires the highest-degree non-seed node with ramping probability; stage two transplants licit-bank features into illicit instances with ramping probability; the harm clock is the stratified-accuracy error with a $0.5/B$ floor (the illicit-class conformal clock is vacuous, $\hat q_1\to1$, reported in Section~\ref{sec:injected}); the headline operating point is the role channel alone at its default threshold, with no tuning on this substrate; two earlier stage-two designs (noise corruption; the vacuous coverage clock) are reported as measured degeneracies, not discarded attempts.

\subsection{Detector and explainer ablations; cost}
\label{app:ablations}
Denominator ablation (pre-registered): under \textsc{remove}, $D_{\text{role}}^{\text{cur}}$ (current-frequency denominator) gives a median lead of $+6.5$ against $+13.0$ for $D_{\text{role}}$ -- conditioning on current frequency quotients out the very support loss under watch. Detectors, each at its own zero-false-alarm threshold: Page--Hinkley $24.5$, CUSUM $23.0$, ADWIN-style window test $27.0$, plain $z$-band $28.5$ (median leads; $10/10$ positive each). Explainers: PGExplainer $24.5$ and gradient-times-adjacency $22.0$ reproduce the lead ($10/10$); a counterfactual edge-deletion score does not ($-7.0$, $0/10$) -- counterfactual scores track prediction flips, which move on the accuracy clock. Cost: one explainer forward pass, $11.8$\,ms per $64$-graph step on CPU ($4.5\times$ a bare prediction), about five times cheaper than degree-MMD in our implementation; the detector update is $O(1)$.

\subsection{The decoy check for presence conditioning}
\label{app:decoy}
Conditioning is meant to separate \emph{the model stopped using this edge} from \emph{this edge disappeared}, so we test it where the two come apart. A third motif $D$ is added to the generator, structurally identical to $A$ and degrading on the same schedule, but carrying no label information: a graph containing $D$ is no more likely to be positive. A statistic that measures the graph will follow $D$; a statistic that measures the model will not. On the fixed-support substrate ($10$ seeds, no evidence jitter, accuracy clock), we rank edges at the alarm step by drop in importance and score precision@$k$ against the truly predictive edges of $A$, with $k$ the size of $A$'s edge set. The conditioned statistic reaches $0.62$; the raw statistic and a degree-based structural ranking both reach $0.00$, their top-$k$ occupied entirely by $D$. The raw mask carries $A_t$ as a factor, so a thinning decoy loses mask mass exactly as a thinning predictive motif does, and nothing in the raw statistic can tell them apart. On the same substrate the two statistics also differ on covariate-only controls, where the graph changes and the model is unharmed: the raw statistic fires in $3\%$ of runs, the conditioned one in $0\%$.

\subsection{Statistical reporting}
Thirty seeds per regime and per control, extended uniformly across all regimes and both control conditions after a five-seed pilot; no regime-specific extension was performed. The scale campaign uses ten seeds per regime at each size. Wilcoxon tests of $L>0$ are Holm-corrected over the family of four regimes predicting a lead; with $n<8$ a one-sided signed-rank test cannot reach $p<2^{-n}$, so we report such cells as undetermined rather than as failures. Leads: medians with interquartile ranges; one-sided Wilcoxon signed-rank tests of $L>0$; detection and false-alarm rates as run fractions; $\Psi$: Spearman correlation of smoothed signal and smoothed coverage gap over the monitored window, per-regime median. Permutation test: $20$ trials of independent per-graph relabelling at a reference step, reporting $|D(\text{perm})-D(\text{orig})|$; bootstrap: $200$ half-batch resamples. Baseline comparison uses per-signal operating points, each calibrated to zero false alarms on its own reference.

\subsection{Lead against transition length}
\label{app:gradient-fig}
\begin{figure}[H]
\centering
\includegraphics[width=0.5\textwidth]{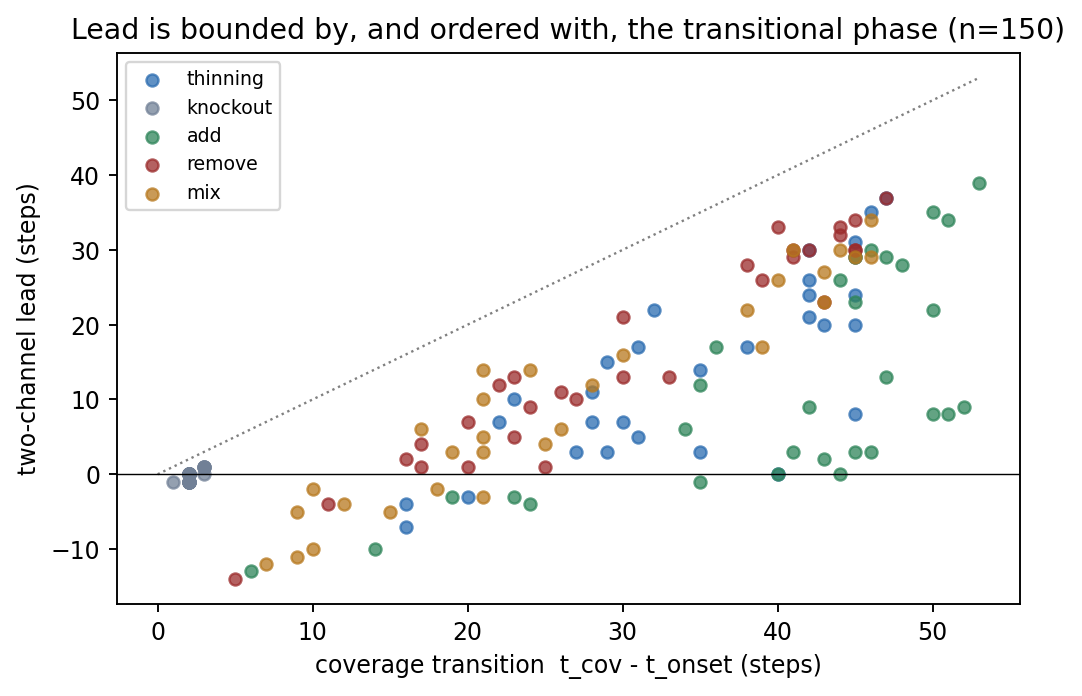}
\caption{All $150$ harmful runs ($30$ seeds per regime): the two-channel lead against the length of the coverage transition. The dotted diagonal is the trivial upper bound $L \le t_{\text{cov}}-t_{\text{onset}}$; Spearman $0.70$ over the five medians and $0.79$ ($95\%$ CI $[0.73, 0.84]$) over the $150$ individual harmful runs; within each regime separately it ranges from $0.66$ to $0.95$.}
\label{fig:gradient}
\end{figure}
Ordered by median transition length, the regimes run \textsc{add} $44$, \textsc{thinning} $35$, \textsc{remove} $30$, \textsc{mix} $22$, \textsc{knockout} $2$ steps; the two-channel leads follow at $+8.5$, $+14.5$, $+13.0$, $+8.0$, $0.0$ (Spearman $0.70$ over the five medians). The one departure from a monotone ordering is \textsc{add}, which is carried by the post-hoc distributional channel rather than by the mass channel the mechanism describes. \textsc{mix} sits low for two reasons: compound shifts spend the conformal margin faster, and its added traffic puts mass on roles shared with the focus set, which can only delay an alarm (\S\ref{sec:signals}).

\subsection{Operating points and the baseline comparison}
\label{app:opbase}
\begin{table}[H]
\centering
\footnotesize
\setlength{\tabcolsep}{5pt}
\begin{tabular}{lrrrrr}
\toprule
Channel & $\lambda$ & pooled $L$ & alarms on \textsc{covariate} & \textsc{thinning} & \textsc{add} \\
\midrule
$D_{\text{role}}$ (mass) & $10$ & $+3.0$ & $0.00$ & $+14$ & $-10$ \\
$D_{\text{role}}^{L1}$ (distribution) & $40$ & $-0.5$ & $0.93$ & $+4$ & $+4$ \\
union & $40$ & $+5.0$ & $0.93$ & $+10$ & $+4$ \\
\midrule
degree MMD & $80$ & $+8.0$ & $1.00$ & $+9$ & $+22$ \\
embedding MMD & $56$ & $0.0$ & $0.97$ & $+13$ & $-13$ \\
\midrule
conditioned identity focus & $24$ & $+10.5$ & $0.00$ & $+22$ & $+25$ \\
\bottomrule
\end{tabular}
\caption{Each channel at the smallest threshold with zero false alarms on the $30$ harmless \textsc{null} runs; pooled median lead over the $150$ harmful runs, then the alarm rate on the $30$ \textsc{covariate} runs, where coverage collapses through background density while the model's reasoning is untouched. A high value there is a correct detection of harm; the attribution channels' $0.00$ is the blind spot of \S\ref{sec:regimes-results}, and what they offer in exchange is that their alarm is specific to a change in the model's reasoning. Leads for two individual regimes are shown for orientation; predictive entropy, the raw identity focus and the full threshold grid are in Figure~\ref{fig:leadfar}, Appendix~\ref{app:leadfar}. The identity channel leads well on this fixed universe and is equally specific -- but a relabelling that changes nothing moves it by $0.69$ (\S\ref{sec:invariance}) and it is undefined under instance sampling (\S\ref{sec:elliptic}). Our claim is not dominance on fixed graphs; it is validity on evolving ones.}
\label{tab:opbase}
\end{table}

\label{app:leadfar}
\begin{figure}[H]
\centering
\includegraphics[width=\textwidth]{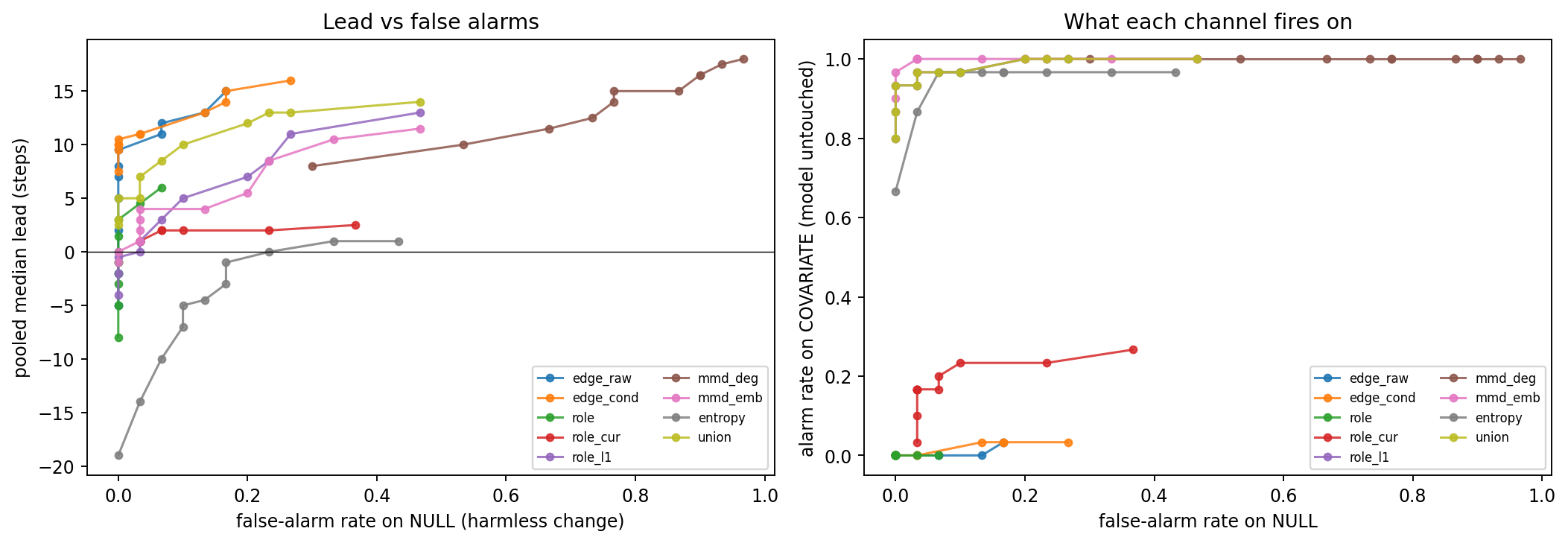}
\caption{Full threshold sweep for every channel. In both panels the horizontal axis is the false-alarm rate on the $30$ harmless \textsc{null} runs as the detector threshold $\lambda$ varies. Left: pooled median lead over the $150$ harmful runs. Right: alarm rate on the $30$ \textsc{covariate} runs, where coverage collapses through background density while the model's reasoning is untouched. The structural baselines detect that harm at every threshold; the attribution channels do not at any threshold, which is the blind spot measured in Table~\ref{tab:main}. Table~\ref{tab:opbase} reads off one point per curve, the smallest $\lambda$ with zero false alarms on \textsc{null}.}
\label{fig:leadfar}
\end{figure}

\subsection{Scale}
\label{app:scale-fig}
\begin{figure}[H]
\centering
\includegraphics[width=\textwidth]{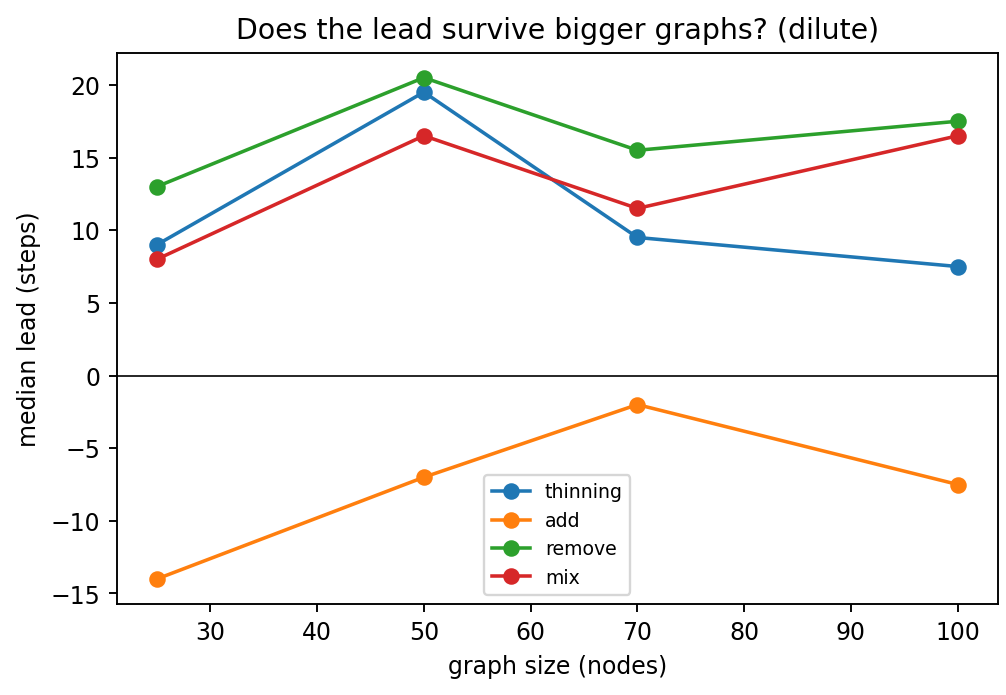}
\caption{Median lead per regime against graph size ($10$ seeds per regime per size). }
\label{fig:scale}
\end{figure}


\subsection{Coverage levels on Elliptic}
\label{app:coverage-levels}
\begin{figure}[H]
\centering
\includegraphics[width=0.85\textwidth]{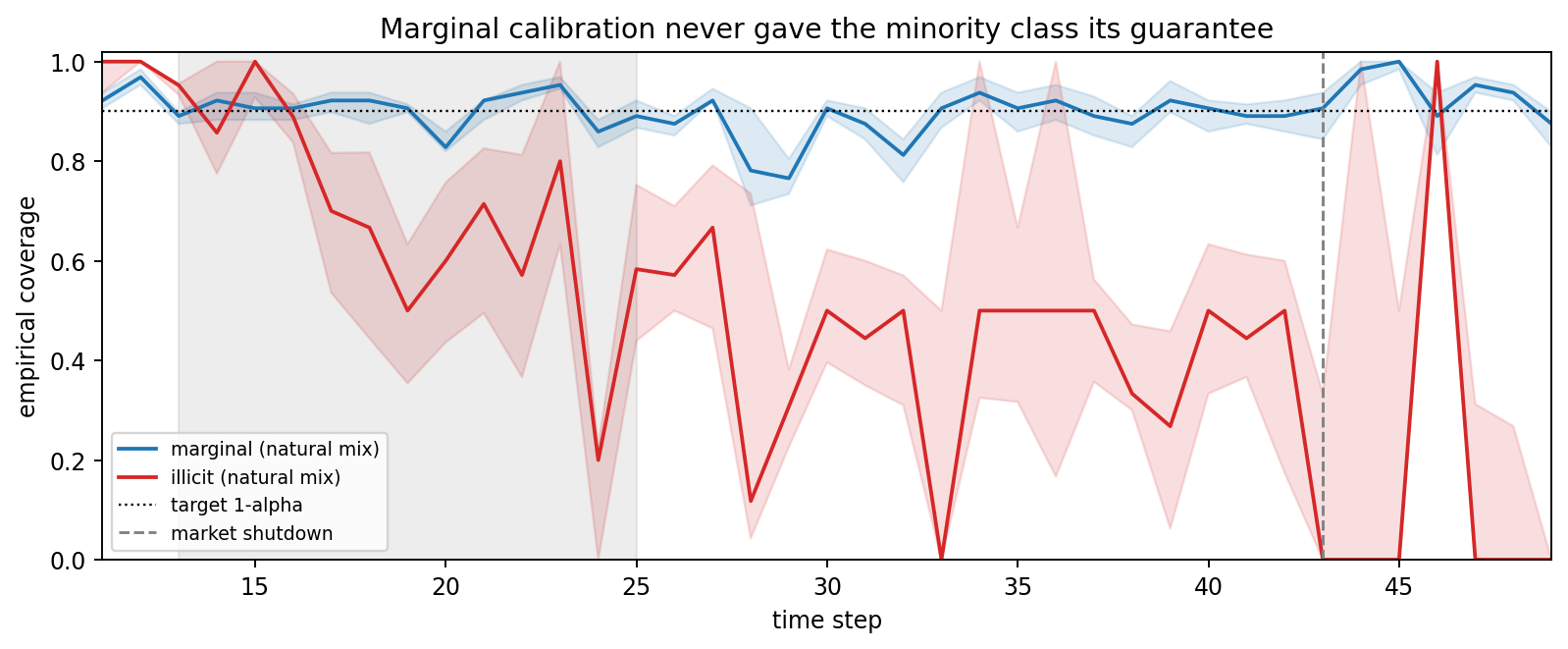}
\caption{Empirical coverage under marginal calibration at the natural $\approx90{:}10$ composition, overall and on the illicit class; median over $15$ repetitions, boxes show the interquartile range. The minority deficit is present on the calibration window itself, widens over the monitored window, and becomes total after the shutdown, while the marginal curve rises above its own calibration level.}
\label{fig:coverage-levels}
\end{figure}
\paragraph{Marginal versus class-conditional coverage.} On the calibration window a single marginal threshold yields $0.901$ coverage overall and $0.681$ on the illicit class (median over $15$ repetitions; paired difference $0.221$, $p=6\cdot10^{-5}$). Over the monitored window illicit coverage falls to $0.459$, and after the shutdown to $0.083$, while the marginal number rises to $0.932$, above its own calibration level; the paired drop differs by $0.572$ ($p=3\cdot10^{-5}$). Under per-class calibration the illicit-class coverage is $0.904$ on the calibration window and collapses in $14/15$ repetitions, a median of three steps before the shutdown.

\paragraph{The vacuous class-conditional clock under injected drift.} The intended harm clock for the injected-drift campaign (\S\ref{sec:injected}) was illicit-class conformal coverage. It turned out vacuous: $\hat q_1\ge0.99$ in $14/15$ repetitions, because a tenth of calibration illicit instances already receive near-zero illicit probability, which pushes the class quantile to the ceiling; at $\hat q_1=1$ the class is never excluded and the guarantee holds by saying nothing. The same pathology appears from the other side: the effective-set-size channel alarms in $11/15$ repetitions, i.e.\ coverage is held by inflating the prediction sets rather than by predicting well. Harm is therefore measured on a stratified-accuracy clock; the conformal remedy (a larger $\alpha_1$) is orthogonal to the question studied here.

\subsection{Notation}
$\mathbf{a}_t$: batch-mean mask vector by node pair; $\varphi_{t,i}$: edge presence; $c_{t,i}$: conditioned importance (Eq.~\ref{eq:cond}); role $\rho$; profile $(m_t(\rho),\varphi_t(\rho))$; $c_t(\rho)$ with reference or current denominator (Eq.~\ref{eq:condrole}); signals $D_{\text{focus}}$, $D_{\text{role}}$, $D_{\text{role}}^{\text{cur}}$, $D_{\text{role}}^{L1}$; coverage gap $g(t)$; onsets $t_{\text{attr}}, t_{\text{acc}}, t_{\text{cov}}$; lead $L$; coupling $\Psi$.

\end{document}